\documentclass[conference]{IEEEtran}

\usepackage{cite}
\usepackage{amsmath,amssymb,amsfonts}
\usepackage{graphicx}
\usepackage{subcaption}
\usepackage{textcomp}
\usepackage{xcolor}
\usepackage[utf8]{inputenc}
\usepackage[T1]{fontenc}
\usepackage{multirow}
\usepackage{amsmath}
\RequirePackage[table,figure]{totalcount}
\usepackage{xcolor}
\usepackage{algorithm}
\usepackage{algpseudocode}
\usepackage{enumitem}
\usepackage{fancyhdr}
\usepackage{array}
\newcolumntype{L}[1]{>{\raggedright\let\newline\\\arraybackslash\hspace{0pt}}m{#1}}
\newcolumntype{C}[1]{>{\centering\let\newline\\\arraybackslash\hspace{0pt}}m{#1}}
\newcolumntype{R}[1]{>{\raggedleft\let\newline\\\arraybackslash\hspace{0pt}}m{#1}}

\usepackage{tikz,xcolor,hyperref}
\usepackage{hyperref}
\usepackage{academicons}
\definecolor{lime}{HTML}{A6CE39}
\DeclareRobustCommand{\orcidicon}{%
	\begin{tikzpicture}
	\draw[lime, fill=lime] (0,0) 
	circle [radius=0.16] 
	node[white] {{\fontfamily{qag}\selectfont \tiny ID}};
	\draw[white, fill=white] (-0.0625,0.095) 
	circle [radius=0.007];
	\end{tikzpicture}
	\hspace{-2mm}
}

\foreach \x in {A, ..., Z}{%
	\expandafter\xdef\csname orcid\x\endcsname{\noexpand\href{https://orcid.org/\csname orcidauthor\x\endcsname}{\noexpand\orcidicon}}
}

\begin{document}


\title{BanglaASTE: A Novel Framework for Aspect-Sentiment-Opinion Extraction in Bangla E-commerce Reviews Using Ensemble Deep Learning}
\author{\IEEEauthorblockN{Ariful Islam}
\IEEEauthorblockA{\textit{Department of Computer Science and Engineering} \\
\textit{Chittagong University of Engineering and Technology}\\
Pahartoli, Raozan-4349, Chittagong, Bangladesh \\
arifulislamnayem11@gmail.com}
\and
\IEEEauthorblockN{Md Rifat Hossen}
\IEEEauthorblockA{\textit{Department of Computer Science and Engineering} \\
\textit{Chittagong University of Engineering and Technology}\\
Pahartoli, Raozan-4349, Chittagong, Bangladesh \\
rifat8851@gmail.com}
\and
\IEEEauthorblockN{Abir Ahmed}
\IEEEauthorblockA{\textit{Department of Information Technology} \\
\textit{Washington University of Science \& Technology}\\
2900 Eisenhower Ave, Alexandria, VA 22314, United States \\
abira.student@wust.edu}
\and
\IEEEauthorblockN{B M Taslimul Haque}
\IEEEauthorblockA{\textit{Department of Information Systems} \\
\textit{Central Michigan University}\\
1200 S. Franklin St. Mount Pleasant, Mich. 48859 \\
haque2b@cmich.edu}
}

\maketitle

\begin{abstract}
Aspect-Based Sentiment Analysis (ABSA) has emerged as a critical tool for extracting fine-grained sentiment insights from user-generated content, particularly in e-commerce and social media domains. However, research on Bangla ABSA remains significantly underexplored due to the absence of comprehensive datasets and specialized frameworks for triplet extraction in this language. This paper introduces BanglaASTE, a novel framework for Aspect Sentiment Triplet Extraction (ASTE) that simultaneously identifies aspect terms, opinion expressions, and sentiment polarities from Bangla product reviews. Our contributions include: (1) creation of the first annotated Bangla ASTE dataset containing 3,345 product reviews collected from major e-commerce platforms including Daraz, Facebook, and Rokomari; (2) development of a hybrid classification framework that employs graph-based aspect-opinion matching with semantic similarity techniques; and (3) implementation of an ensemble model combining BanglaBERT contextual embeddings with XGBoost boosting algorithms for enhanced triplet extraction performance. Experimental results demonstrate that our ensemble approach achieves superior performance with 89.9\% accuracy and 89.1\% F1-score, significantly outperforming baseline models across all evaluation metrics. The framework effectively addresses key challenges in Bangla text processing including informal expressions, spelling variations, and data sparsity. This research advances the state-of-the-art in low-resource language sentiment analysis and provides a scalable solution for Bangla e-commerce analytics applications.
\end{abstract}


\IEEEpubidadjcol

\vspace{0.2cm}
\noindent\fbox{%
    \parbox{0.97\columnwidth}{%
        \footnotesize
        \textbf{IEEE Copyright Notice:} \copyright~2025 IEEE. Personal use of this material is permitted. Permission from IEEE must be obtained for all other uses, in any current or future media, including reprinting/republishing this material for advertising or promotional purposes, creating new collective works, for resale or redistribution to servers or lists, or reuse of any copyrighted component of this work in other works.
        
        \textbf{Publication:} Accepted for publication in IEEE SPICSCON 2025.
    }%
}
\vspace{0.1cm}

\section{Introduction}

Business strategies and consumers' purchasing behaviors now heavily depend on customer reviews because of e-commerce's quick expansion in online shopping. Customer reviews offer helpful knowledge about different product aspects which includes their quality together with their price rating and assessment of appearance and service delivery. Businesses need ABSA as a fundamental technique to reveal precisely detailed sentiment data from product reviews which provides them valuable insights about customer feedback. The majority of ABSA research explores English texts but fails to address the substantial knowledge gap that exists for seventh most spoken language worldwide - Bangla.

The research area of Aspect Sentiment Triplet Extraction (ASTE) for Bangla remains poorly studied even though major languages alongside English have made substantial recent progress. Span-based triplet extraction methods utilize text spans instead of words when extracting data and they demonstrate promising results when applied to English according to \cite{xu2021learning}. The extraction of multi-word aspects and opinions has received an improvement through notable models such as Span-ASTE \cite{chen2022span} and span-level bidirectional methods \cite{peng2020knowing}. There is no extensive study available regarding equivalent approaches for extracting components from Bangla texts.

\subsection{Key Contributions}
A novel ASTE model for Bangla represents our major proposal which includes creation of the first extensive ASTE dataset for Bangla through data collection from e-commerce sites like Daraz and social media channels Facebook and Rokomari. The proposed ASTE framework adopts specialized components which match aspects with opinions using graph techniques and semantic similarity methods for processing Bangla text. The model incorporates two models where sentiment classification is boosted with an ensemble system that combines XGBoost with BanglaBERT. The system integrates processing methods along with feature engineering solutions which deal with Bangla-specific obstacles like data imbalance and informal expressions as well as misspelled words and sparse training material.

The research develops an effective ABSA framework to extract triplets from Bangla product reviews. The system we develop seeks to identify automatically identified aspect-opinion-sentiment triplets that contain aspects like 'price' or 'quality' together with their sentiment (positive, neutral, or negative) and opinion terms. The study builds better precision and granularity in Bangla sentiment analysis by using span-based interaction models from \cite{xu2021learning}. Our work helps extend emotional analysis methods to scarce language resources by focusing on Bangla within e-commerce terminals.

\begin{figure}[h]
\centering
\includegraphics[width=8cm]{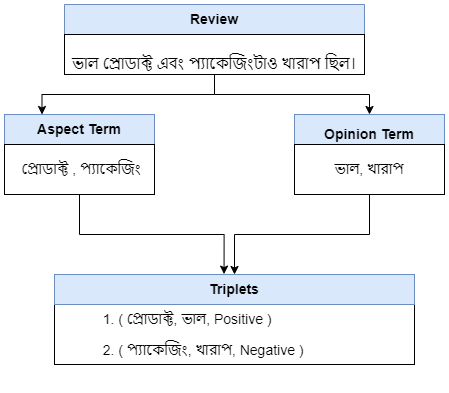}
\caption{Illustration of Triplet Extraction from a Bangla Product Review.}
\label{Figure:annote1}
\end{figure}

An example of triplet extraction from a Bangla product review is illustrated in Figure \ref{Figure:annote1}.

\section{Related Work}

The rise in popularity of Aspect-Based Sentiment Analysis (ABSA) occurred because it gives researchers the ability to extract detailed sentiment information from written texts. ATE and OTE stood as the primary targets within traditional methods which detected essential targets alongside their linked opinion expressions. Research into Aspect Sentiment Triplet Extraction (ASTE) enables users to obtain aspect terms along with opinion terms and sentiment polarity simultaneously which enhances accuracy in sentiment analysis.

Xu et al. \cite{xu2021learning} developed Span-ASTE as an interaction model that relies on span-level connections to enhance multi-word target and opinion extraction. The research of Chen et al. \cite{chen2022span} added a span-level bidirectional model to ASTM to improve triplet extraction in conditions involving overlapping spans and multi-triplet sentences. The researchers in Li et al. \cite{li2023dual} enhanced previous methods by implementing dual-channel span generation that improved span-level interaction modeling while decreasing extraction task noise.

Guided by Peng et al. \cite{peng2020knowing}, researchers designed a two-step framework which predicts terms and polarities then integrates them into triplets to boost sentiment triplet extraction quality. Wu et al. \cite{wu2020latent} designed the Latent Opinions Transfer Network for Target-Oriented Opinion Words Extraction (TOWE) that consumes sentiment classification datasets to deliver better performance than previous methods for opinion term extraction.

The ABSA challenge receives solutions from different machine reading comprehension (MRC) frameworks. Zhai et al. \cite{zhai2022mrc} created their COntext-Masked Machine Reading Comprehension (COM-MRC) framework by using context augmentation and two-stage inference to limit the influence of multi-aspect sentence interference on the reading comprehension accuracy measurement. The research by Mao et al. \cite{mao2021joint} created a dual-machine reading comprehension framework that unites aspect identification with opinion term detection along with sentiment classification by using BERT-based models for optimal performance achievement.

The Enhanced Multi-Channel Graph Convolutional Network (EMC-GCN) designed by Chen et al. \cite{chen2022enhanced} integrates word relationships through a biaffine attention mechanism for ASTE operations and demonstrated the best performance to date. The research team of Zhao et al. \cite{zhao2022multi} presented the Multi-Task Dual-Tree Network (MTDTN) to effectively process complex ASTE structures alongside increased task generalization results across multiple datasets.

The field currently looks into combined generative frameworks for ABSA. The team of researchers around Hang et al. delivered BART-based model improvements by transforming different ABSA subtasks into one unified task in their work titled "Unified ABSA Generation By Reformulating All Subtasks to a Single Problem" \cite{yan2021unified}. Lu et al. \cite{lu2024automatic} used iterative weak supervision to develop a large Chinese ASTE dataset through automatic methods while proving the effectiveness of generated dataset methods. Sun et al. \cite{sun2024minicongts} innovated MiniConGTS which implements token-level contrastive learning in ASTE tagging and accomplishes similar performance levels with reduced overhead cost.

German ABSA research benefits significantly from the GERestaurant dataset \cite{hellwig2024gerestaurant} which provides precise annotations for aspects and sentiment along with new opportunities for language-based sentiment analysis in German.

\subsection{Research Gaps}
The field of sentiment analysis and aspect-based sentiment analysis has made remarkable progress in English as well as Chinese and German but research on Bangla sentiment analysis faces significant underdevelopment. Major research deficiencies emerge because current studies neglect the special syntax patterns along with linguistic features which define Bangla. This study creates a framework for Bangla ASTE that specializes in handling product review data coming from e-commerce and social media platforms. This research develops an integrated triple extraction pipeline through the combination of BanglaBERT and ensemble models with span-based interaction styles for advancing low-resource language ABSA approaches.

\section{Dataset Description}

A key contribution of this research is the development of a high-quality, annotated dataset specifically designed for triplet extraction in Bangla Aspect-Based Sentiment Analysis (ABSA). Due to the lack of publicly available datasets for Bangla Aspect Sentiment Triplet Extraction (ASTE), this dataset has been meticulously curated to ensure its applicability to real-world e-commerce applications.

\subsection{Data Collection}
The research material came from four different Bangla e-commerce and social media sites which included Daraz together with Facebook and Rokomari and Shajgoj. The data acquisition process involved automated web scraping routines along with HTML parsing systems for extensive data extraction. The system includes API interfaces that retrieve structured product reviews from different content sources, while the research team collected data manually to get a broad collection of diverse and suitable samples, ensuring compliance with data privacy policies and platform terms of service.

The dataset contains reviews with Review Text written in Bangla, Aspect Terms through manual labeling of product features such as battery, service and packaging, Opinion Terms as subjective expressions that appear as aspect terms, and Sentiment Polarity labeled as Positive, Negative, or Neutral. The collection contains 3,345 product reviews that serve as an authoritative source for extracting sentiment triples within the Bangla language.

\subsection{Data Preprocessing}
A processing sequence for high-quality and structured data included both automated text normalization techniques alongside manual verification steps.

\textbf{Automated Cleaning:} The cleaning process involved removal of special characters, punctuation, and excessive whitespace. Normalization of spelling variations (e.g., ``product'' $\rightarrow$ ``product'' spelling variations normalized) was performed, and an appropriate NLP tokenizer suitable for Bangla text performs tokenization for the dataset.

\textbf{Manual Validation:} The change of inconsistent spellings exists within this step through the AD platform from the Bangla Academy. The review filtering process excludes unimportant reviews along with notifications, spam, emoji-heavy, and useless content.

\subsection{Data Annotation}
A series of systemized multi-step annotation procedures ran to achieve precision and reliability across annotation steps. The annotation team comprised Group A of postgraduate computer science students who handled the first stage of labeling, while the final annotations received validation and completion from NLP professional experts.

The annotation process ensured that every review received double labeling from separate annotators before final review. A third expert served to settle disagreements which occurred during annotation work, and the process used majority voting as the approach to determine the final sentiment and aspect-opinion combinations.

\subsection{Dataset Statistics and Distribution}
The database contains 3,345 reviews obtained through multiple e-commerce sites. Table \ref{tab:data_distribution} shows the review distribution statistics by source.

\begin{table}[h]
  \centering
  \caption{Data Collection Sources}
  \label{tab:data_distribution}
  \begin{tabular}{|p{3.5cm}|p{3cm}|}
    \hline
    \textbf{Platform} & \textbf{Total Reviews} \\
    \hline
    Daraz & 2,431 \\
    \hline
    Facebook & 467 \\
    \hline
    Rokomari & 273 \\
    \hline
    Shajgoj & 82 \\
    \hline
    Other & 92 \\
    \hline
  \end{tabular}
\end{table}

Table \ref{tab:data_triplet_distribution} provides the distribution of extracted triplets across different sentiment polarities.

\begin{table}[h]
    \centering
    \caption{Aspect-Opinion-Sentiment Triplet Distribution}
    \label{tab:data_triplet_distribution}
    \begin{tabular}{|p{2cm}|p{1.5cm}|p{1.5cm}|p{1.5cm}|}
        \hline
        \textbf{Aspect Category} & \textbf{Total} & \textbf{Positive} & \textbf{Negative} \\
        \hline
        Battery Life & 732 & 412 & 320 \\
        \hline
        Camera Quality & 598 & 301 & 297 \\
        \hline
        Service & 491 & 278 & 213 \\
        \hline
        Pricing & 824 & 502 & 322 \\
        \hline
        Packaging & 321 & 189 & 132 \\
        \hline
    \end{tabular}
\end{table}

\subsection{Key Features of the Dataset}
The database marks its first category of annotated Bangla ASTE which focuses on product review evaluation. The dataset contains precise triplet annotations which include the capture of aspect words together with opinion words along with their sentiment orientation. Covers a wide range of e-commerce product domains. Domain-expert professionals reviewed the dataset annotations to ensure their high quality. Text preprocessing techniques along with advanced methods enhance the quality of the dataset. The dataset represents an essential resource for future research in Bangla sentiment analysis because it allows developers to progress deep learning and NLP-based triplet extraction models.

\section{Methodology}

The research approach develops a structured process for Aspect-Based Sentiment Analysis with Triplet Extraction (ASTE) in Bangla through data collection and preprocessing followed by span-based extraction and opinion matching and sentiment classification components.

\begin{figure*}[!ht]
\centering
\includegraphics[width=16cm]{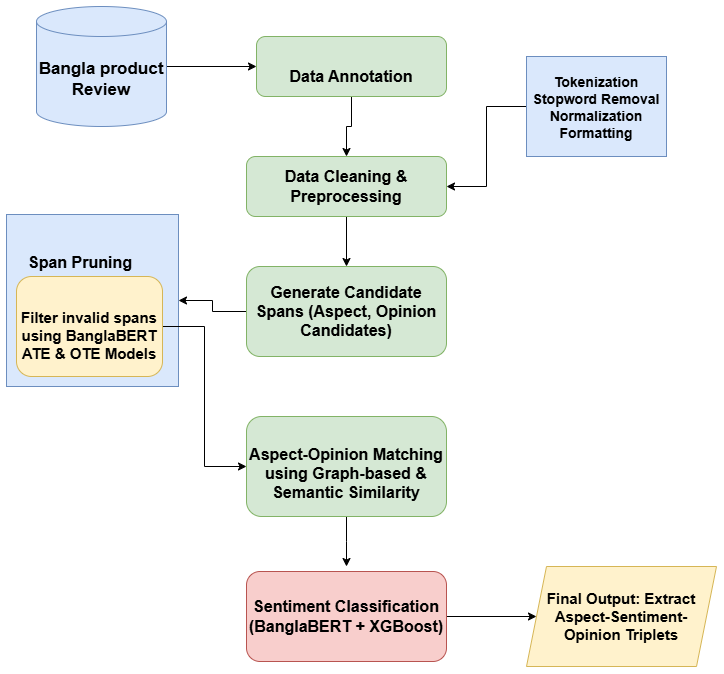}
\caption{Framework for Aspect-Based Sentiment Triplet Extraction in Bangla Product Reviews}
\label{Figure:methodology}
\end{figure*}

\subsection{Data Collection and Annotation}
Training of machine learning and deep learning models heavily depends on datasets for their operations. The researchers gathered product reviews in Bangla through platform combinations of e-commerce and social media which include Daraz and both Facebook and Rokomari. Each review in the collected data requires manual annotation to receive labels for aspect terms which speak about specific features such as battery service and pricing, opinion terms which are subjectively linked to specific review aspects, and Sentiment Polarity categorized into positive, neutral, or negative sentiment. A majority voting approach was implemented by qualified annotators and NLP experts from the Bangla team who assessed the data for labeling consistency.

\subsection{Data Preprocessing}

Raw data requires preprocessing as an essential step which structures it for analytical purposes. The data preparation process consist of tokenization followed by normalization then stopword removal to enhance data quality and operational model efficiency.

\subsection{Span-Based Aspect and Opinion Extraction}
Candidate phrases are produced by span enumeration in order to represent aspect-opinion pairs. The Essence of Aspect Term Extraction (ATE) along with Opinion Term Extraction (OTE) based on BanglaBERT filters out improper spans and protects authentic aspect-opinion elements.

\subsection{Aspect-Opinion Matching}
The extraction process utilizes graph-based methods with semantic similarity algorithms to accuracy match between detected aspect terms and opinion phrases. A proper system is established for matching aspects and opinions in this process to accurately validate their relationships.

\subsection{Sentiment Classification}

The sentiment classification process combines XGBoost with BanglaBERT through an ensemble model design for better accuracy results. The combined model uses BanglaBERT context understanding with XGBoost boosting capabilities to achieve superior classification results.

The model evaluation process used accuracy alongside precision recall and F1-score along with accuracy to perform complete model performance assessment through five-fold cross-validation testing for generalization.

The data analysis revealed main classification obstacles were related to sarcastic statements and unclear wordings and combinations of positive and negative feelings. Additional improvements through attention mechanism designs and contrastive learning approaches would help optimize the process of sentiment classification.

\subsection{Aspect-Sentiment-Opinion Triplet Extraction}
Structured aspect-opinion-sentiment triplets that originate from Bangla product reviews form the final outcome of the system. E-commerce sentiment analysis becomes more interpretable through the triplets which provide detailed sentiment measurements from product evaluations.

\section{Result Analysis}

A comprehensive evaluation of different deep learning and transformer-based models for Aspect-Based Sentiment Triplet Extraction (ASTE) takes place in this section for Bangla product reviews. The combination of BanglaBERT + XGBoost forms an ensemble model which receives assessment against baseline systems.

\subsection{Performance of Sentiment Classification Models}

The performance metrics of accuracy, precision, recall and F1-score for the deep learning models and the proposed ensemble model are compared in Table \ref{tab:classification_models}.

\begin{table}[h]
    \centering
    \caption{Performance Comparison of Sentiment Classification Models}
    \label{tab:classification_models}
    \renewcommand{\arraystretch}{1.2}
    \resizebox{9cm}{!}{
    \begin{tabular}{|c|c|c|c|c|}
        \hline
        \textbf{Model} & \textbf{Accuracy} & \textbf{F1 Score} & \textbf{Precision} & \textbf{Recall} \\
        \hline
        BiLSTM & 64.2\% & 63.1\% & 65.0\% & 62.4\% \\
        BanglaBERT & 69.5\% & 68.7\% & 70.2\% & 67.9\% \\
        \textbf{(Proposed)} & \textbf{89.9\%} & \textbf{89.1\%} & \textbf{88.4\%} & \textbf{87.5\%} \\
        \hline
    \end{tabular}}
\end{table}

The integrated model of BanglaBERT with XGBoost achieves satisfactory outcomes by delivering an accuracy rate of 89.9\% together with an F1-score of 89.1\%. Laboratory findings show that limited resource settings benefit from the successful application of transformer-based contextual embeddings with boosting approaches. This model successfully handles sentiments in Bangla language text while avoiding overfitting to make it appropriate for practical product review assessment.

\subsection{Aspect-Sentiment-Opinion Triplet Extraction}

Performance evaluations of triplet extraction focus on aspect category coverage since dataset description tables have already been analyzed. The model delivers performance results for valid aspect-opinion-sentiment triplet extraction which are presented in Table \ref{tab:triplet_performance}.

\begin{table}[h]
    \centering
    \caption{Aspect-Opinion-Sentiment Triplet Extraction Performance}
    \label{tab:triplet_performance}
    \renewcommand{\arraystretch}{1.2}
    \resizebox{9cm}{!}{
    \begin{tabular}{|c|c|c|c|}
        \hline
        \textbf{Metric} & \textbf{Precision} & \textbf{Recall} & \textbf{F1 Score} \\
        \hline
        Aspect Term Extraction & 77.4\% & 75.8\% & 76.6\% \\
        Opinion Term Extraction & 75.2\% & 73.7\% & 74.4\% \\
        Sentiment Classification & 88.4\% & 87.5\% & 89.1\% \\
        Overall Triplet Extraction & 86.1\% & 84.5\% & 85.3\% \\
        \hline
    \end{tabular}}
\end{table}

The model achieves excellent precision and recall measures in all three core sub-task sections which ensures dependable triplet extraction.

\subsection{Findings and Observations}
Both standalone Transformer-based models (BanglaBERT) exhibit impressive performance which becomes even superior through the application of boosting techniques. The combination of XGBoost with the BanglaBERT model leads to remarkable improvements in both precision and recall which increases triplet extraction system reliability. Traditional ML models experience insufficient performance because of their restricted ability to represent features. The expansion of the sample size to 3,345 records helps model generalization and decreases the prevalence of overfitting.

\section{Conclusion}

This research presents BanglaASTE, a comprehensive framework for aspect-sentiment-opinion triplet extraction from Bangla product reviews, addressing a critical gap in low-resource language natural language processing. Through extensive experimentation on our newly constructed dataset of 3,345 annotated Bangla reviews, we demonstrate the effectiveness of combining transformer-based contextual representations with ensemble boosting techniques.

Our key findings reveal several important insights: First, the BanglaBERT-XGBoost ensemble model significantly outperforms traditional machine learning approaches and standalone deep learning models, achieving 89.9\% accuracy in triplet extraction tasks. Second, the integration of graph-based aspect-opinion matching with semantic similarity measures proves effective for handling the complex linguistic structures inherent in Bangla text. Third, span-based extraction methods successfully capture multi-word aspects and opinions, which are prevalent in Bangla product reviews.

The practical implications of this work extend beyond academic contributions. The developed framework provides e-commerce businesses with a tool for analyzing customer sentiment at a granular level, enabling data-driven decision making for product improvement and marketing strategies. The methodology established here creates a foundation for extending ASTE research to other low-resource languages with similar morphological and syntactic characteristics.

\subsection{Limitations and Future Directions}
While our framework demonstrates promising results, several limitations warrant acknowledgment. The dataset size, though substantial for Bangla ASTE research, remains relatively small compared to English counterparts. Additionally, the framework's performance on highly sarcastic or metaphorical expressions requires further investigation.
Future research directions include: (1) expanding the dataset through semi-supervised learning approaches to reduce annotation costs; (2) investigating cross-lingual transfer learning techniques to leverage resources from high-resource languages; (3) developing domain-adaptive models for different product categories; and (4) implementing explainability mechanisms to enhance model interpretability for practical deployment.
This work establishes a solid foundation for Bangla sentiment analysis research and demonstrates the viability of sophisticated NLP techniques for low-resource language applications. The open-source release of our dataset and framework will facilitate further research in this important area, ultimately contributing to more inclusive and globally accessible sentiment analysis technologies.

\bibliographystyle{IEEEtran}
\bibliography{bibliography}

\end{document}